\documentclass[letterpaper, 10 pt, conference]{ieeeconf}  

\IEEEoverridecommandlockouts                              

\title{\LARGE \bf
SelfDeco: Self-Supervised Monocular Depth Completion in Challenging Indoor Environments 
}

\author{Jaehoon Choi$^{1,2}$, Dongki Jung$^{1}$, Yonghan Lee$^{1}$, Deokhwa Kim$^{1}$, Dinesh Manocha$^{2}$, and Donghwan Lee$^{1}$\\
$^{1}$NAVER LABS $^{2}$University of Maryland
}

\usepackage{graphicx}
\usepackage{amsmath}
\usepackage{amssymb}
\usepackage{multirow}
\usepackage{booktabs}

\begin{document}
\maketitle
\thispagestyle{empty}
\pagestyle{empty}

\begin{abstract}
We present a novel algorithm for self-supervised monocular depth completion. Our approach is based on training a neural network that requires only sparse depth measurements and corresponding monocular video sequences without dense depth labels.
Our self-supervised algorithm is designed for challenging indoor environments with textureless regions, glossy and transparent surfaces, moving people, longer and diverse depth ranges and scenes captured by complex ego-motions. Our novel architecture leverages both deep stacks of sparse convolution blocks to extract sparse depth features and pixel-adaptive convolutions to fuse image and depth features. We compare with existing approaches in NYUv2, KITTI and NAVERLABS indoor datasets, and observe 5\:-\:34 \% improvements in root-means-square error (RMSE) reduction.
\end{abstract}

\section{Introduction}


Depth completion has been widely studied in the field of robot navigation, computer vision, and autonomous driving. Its goal is to convert a sparse depth map from active depth sensors such as LiDAR or RGB-D cameras to a dense depth map. 
In practice, robots with accurate sensing and dense depth maps have a lower chance of collision \cite{dinesh1} and are better able to compute safe navigation routes through challenging environments \cite{dinesh2,dinesh3}. Currently, most recent methods for depth completion rely on supervised learning \cite{CSPN++,2020Eldesokey,GuideNet}, requiring large amounts of high-quality and dense depth maps as ground truth.
Unfortunately, generating dense depth maps is expensive and challenging due to the sparse and noisy depth values from other active sensors. Even existing expensive 3D LiDAR sensors only have a limited number of scan lines and provide sparse depth measurements. 

Self-supervised methods with monocular videos that are able to train depth completion neural networks without the need for dense depth maps. These methods \cite{2017Godard,SfMLearner,2018DDVO,Monodepth2} are based on an unsupervised learning framework that use depth and camera ego-motion from monocular videos and use photometric loss as supervision. This loss measures the difference between a reference image and a synthesized image obtained by the depth-guided reprojection of other views into the reference view. However, the previous self-supervised training process for depth completion \cite{2019ICRAMa, 2019CVPRYang} may not work well on public indoor datasets such as NYUv2 \cite{NYUv2}. This is due to large areas of textureless regions and scenes captured by complex ego-motions of a handheld camera \cite{MovingIndoor,weakrect}.
Large indoor environments are more challenging not only because of  the above issues but also because of large regions of glossy and transparent surface,
non-Lambertian surfaces, longer and diverse ranges, and moving people. To overcome these issues, our proposed training process requires effective ways to compute complex camera ego-motions attached to a robot, mask out regions where photometric loss is unreliable, and output depth maps robust to diverse depth range. In this paper, we address the problem of developing self-supervised depth completion methods that work robustly on general indoor environments.  

 \noindent\textbf{Main Results:} We improve the training framework by leveraging learning-based local features to obtain accurate relative poses, making robust matching with auto-masking and minimum reprojection loss \cite{Monodepth2}, and predicting inverse depth using a sigmoid function. However, we observe that merely masking out all regions still leads to depth artifacts including blurred boundaries and irregular patterns copied from sparse depth input. Therefore, we also propose a novel architecture with two unique components: sparsity-aware convolutions that extract features properly from sparse input and pixel-adaptive convolutions which propagate this information to the whole depth prediction with the guidance of image features. We highlight the effectiveness of our method by evaluating our approach on challenging datasets such as the Metro Station or the Department Store (Fig. \ref{intro_figure}). We also demonstrate the applicability of our method on datasets for which prior self-supervised depth estimation and completion fail to provide reliable depth maps. \textbf{Our novel contributions} can be summarized as follows: 

\begin{figure}[t]
    \centering
    \includegraphics[width=0.87\linewidth]{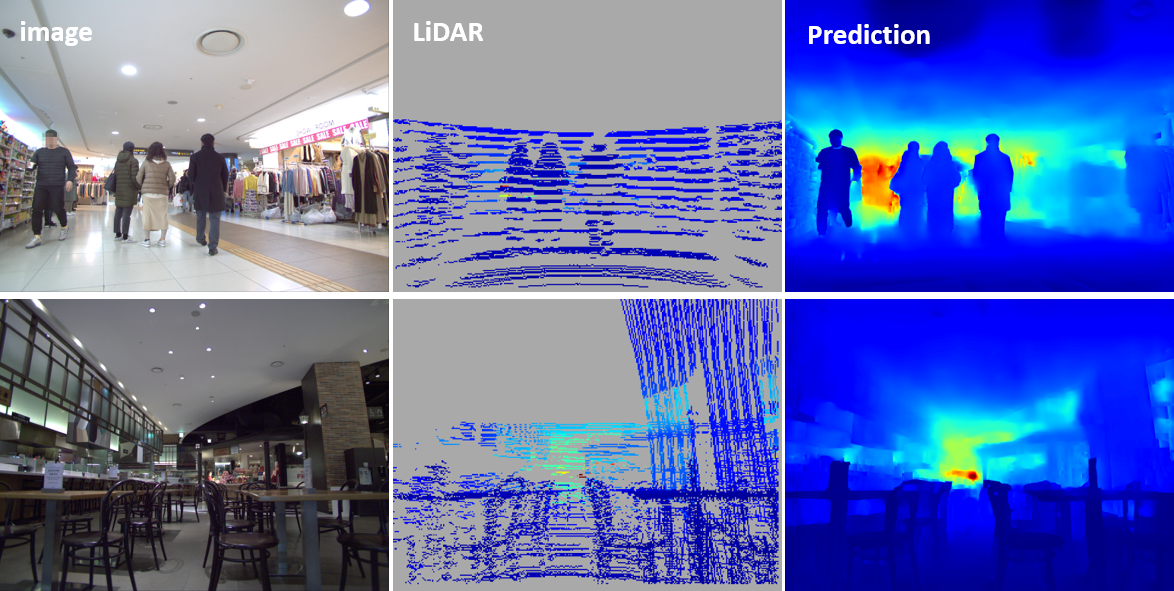}
    \vspace{-2mm}
    \caption{Depth completion results in challenging scenes: Metro Station (top) and the Department Store (bottom) from NAVERLABS indoor dataset. Our method takes an RGB image and a sparse depth map (projection of LiDAR pointcloud) as input, and predicts a dense depth map. 
    (low \protect\includegraphics[width=0.5cm,height=0.2cm]{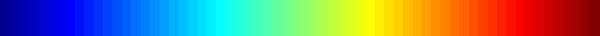} high; grey means empty depth values.)}
    \label{intro_figure}
    \vspace{-4mm}
\end{figure}

\begin{itemize}
    \item We develop a robust and stable training framework for self-supervised depth completion in challenging indoor environments.
    \item We propose a novel architecture that can extract sparse depth features guided by high-level image features.
    \item Our method outperforms prior self- and fully- supervised methods on challenging indoor datasets like NAVERLABS. We also achieve the state-of-the-art performance among self-supervised methods on public benchmark datasets, including KITTI and the NYUv2 dataset based on various evaluation metrics.
\end{itemize}


\section{Related Work}
Recently, depth completion methods using a deep neural network have shown promising results. Uhrig et al. \cite{SparsityCNN} propose a sparsity-invariance convolution layer to handle sparse depth or features by using mask normalizations. Eldesokey et al. \cite{2018Eldesokey,2019Eldesokey} propose the normalized convolution to produce both a confidence map and dense depth output. In \cite{2020Eldesokey}, they further extend their work into uncertainty estimation with depth completion methods. Teixeira et al. \cite{Aerial} use this confidence propagation from \cite{2018Eldesokey} to compute dense depth from LiDAR. 
Cheng et al. \cite{2018ECCVcspn,CSPN++} utilize the neural network to learn the affinity among neighboring pixels in order to perform the propagation process. In addition to using RGB images, some methods take advantage of information from other modalities, e.g., surface normals \cite{2018Funkhouser,deeplidar,depthnormal} or segmentation \cite{2018segmentation}. Yang et al. \cite{2019CVPRYang} predict the posterior over depth maps with a conditional prior network. Some works \cite{2019ICCVWIndoor,2020BMVCDecoder} only focus on an indoor depth completion task. However, all aforementioned works on indoor scenarios are limited to a small-scale indoor environment 
and handle non-LiDAR indoor data (e.g., on the NYU-Depth-v2 \cite{NYUv2} and Matterport3D \cite{2018Funkhouser} datasets). In this paper, we introduce challenging indoor scenarios such as the department store or metro station containing LiDAR sensor data. 

The lack of dense depth maps is a significant limitation of supervised depth completion. As an alternative, several current works study self-supervised depth completion methods.
Ma et al. \cite{2019ICRAMa} present self-supervised methods based on a photometric loss with pose estimation using the PnP method. Yang et al. \cite{2019CVPRYang} present a conditional prior network (CPN) that predicts a probability of depths at each pixel. Zhang et al. \cite{Dfinenet} propose an end-to-end framework to jointly estimate pose and depth without the PnP method. Yoon et al. \cite{2020IROSYoon} combine monocular depth estimation and Gaussian process-based depth regression for depth completion. 


\section{Proposed Method} 
\subsection{Training Framework}
\begin{figure}[t]
\centering
\includegraphics[width=0.9\linewidth]{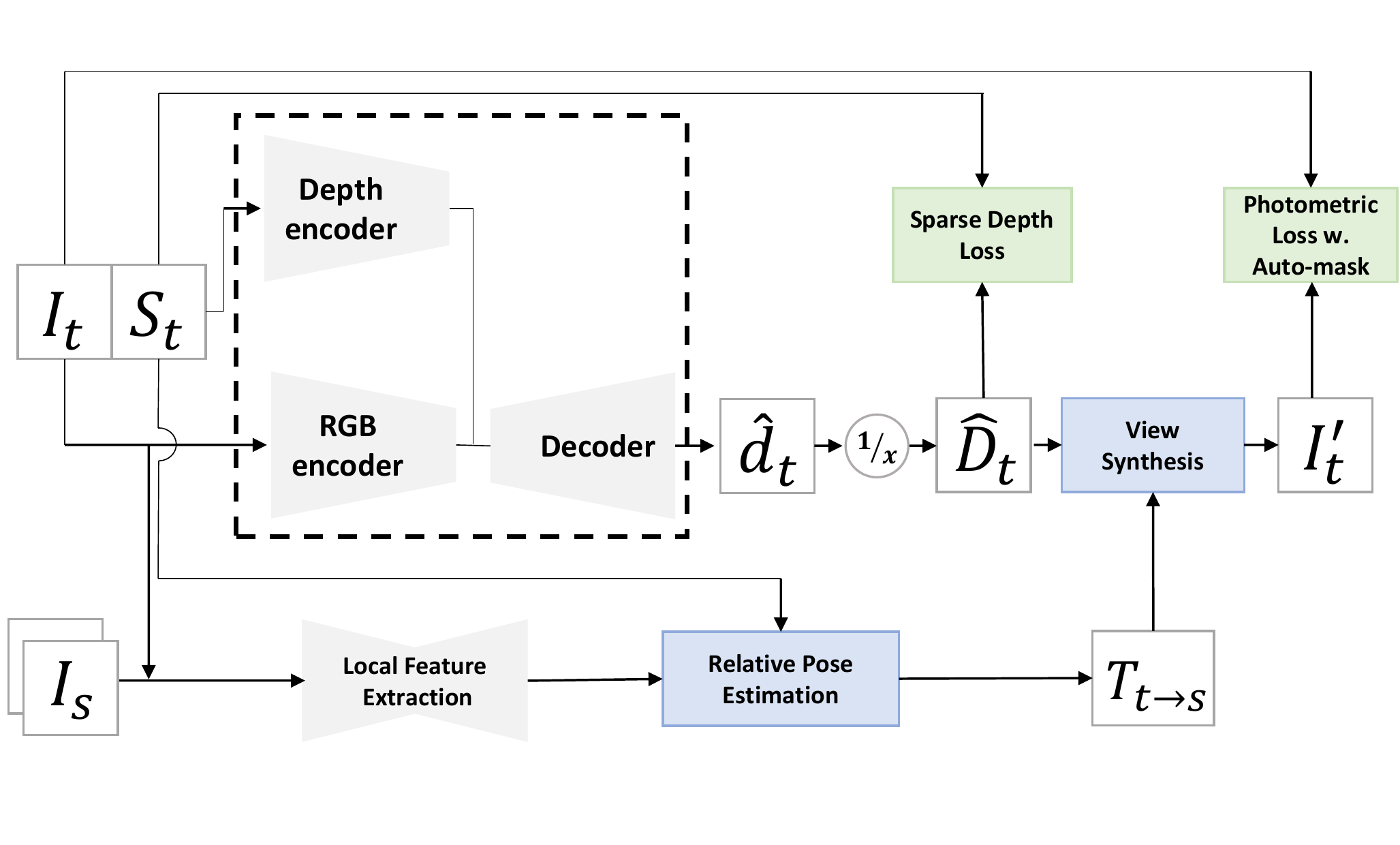}
\vspace{-6mm}
\caption{Overview of our proposed framework.
The black block with a dash-dotted line shows the depth completion network (details in Fig. \ref{figure2}). 
}
\vspace{-4mm}
\label{figure1}
\end{figure}

In Fig. \ref{figure1}, our proposed depth completion model takes a target view image \(I_{t}\) and a sparse depth \(S_{t}\) as input, and outputs a dense inverse-depth \(\hat{d}_{t}\). We invert \(\hat{d}_{t}\) to compute a dense depth map \(\hat{D}_{t}\). In our method, the source view images contains its two adjacent temporal frames, i.e., \(I_{s} \in \{I_{t-1}, I_{t+1}\}\), although including a larger temporal context is possible. Also, we estimate the relative pose \(T_{t\rightarrow s}\) between the target image \(I_{t}\) and the set of source images \(I_{s}\) by using a Perspective-n-Point (PnP) \cite{PnP} with Random Sample Consensus (RANSAC) \cite{RANSAC}.  

\subsubsection{Relative Pose Estimation}
Although most prior works \cite{Monodepth2,Dfinenet} adopt learning-based pose estimation, we observe that learning-based pose estimation reveals degradation in performance when applied to challenging environments such as indoor scenes \cite{MovingIndoor,towardsbetter,weakrect}. Additionally, according to recent findings \cite{sattlerlearningpose}, learning-based pose estimation methods suffer from scale ambiguity problems and make it difficult to provide absolute-scale depth supervision. Unlike self-supervised depth estimation, pose estimation via the PnP method is aligned with the depth completion task because our model needs to predict absolute scale depth. 

Our method first detects learning-based features and computes their corresponding descriptors via learning-based algorithms. We select R2D2 \cite{r2d2}, which shows better performance than traditional hand-crafted features on the visual localization benchmarks. 
Given the predicted feature in source and target images, we can compute high-quality correspondences by using a FLANN-based search algorithm \cite{FLANN} and then output the relative camera pose by solving PnP with corresponding sparse depths. 
Compared to hand-crafted features used by \cite{2019ICRAMa}, dense correspondences obtained by R2D2 enable us to fully utilize sparse depth measurements with corresponding feature points for PnP pose estimation. 

\subsubsection{Photometric Loss}
Given the camera intrinsic matrix \(K\) and the relative pose \(T_{t\rightarrow s}\), we compute the warped pixel coordinates and synthesize target image \(I^\prime_{t}\) from source images \(I_{s}\) via a bilinear sampling function. Following the self-supervised depth estimation \cite{2017Godard,SfMLearner}, we use the combination of the L1 and SSIM \cite{SSIM} as the photometric loss \(L_{ph}\): 

\begin{equation}
\begin{split}
    L_{ph}(I_{t}, I^\prime_{t}) &= \alpha \frac{1-\text{SSIM}(I_{t}, I^\prime_{t})}{2} 
    + (1-\alpha)\parallel I_{t}-I^\prime_{t} \parallel, \\
    I^\prime_{t}(x) &= I_{s}\langle \pi( KT_{t\rightarrow s}\hat{D}_{t}(x)K^{-1}x_{h}) \rangle\:.
\end{split}
\label{photometric_loss}
\end{equation}

where \(x_{h} = [x^{T}, 1]^{T}\) are the homogeneous coordinates from the target image, \(\pi\) gives pixel coordinates,  and \(\langle \cdot \rangle\) denotes the bilinear sampling function. We adopt the minimum reprojection loss and auto-masking from \cite{Monodepth2}. Our methods often face the occlusion issue, meaning that some pixels from frame \(I_{t}\) do not appear in either \(I_{t-1}\) or \(I_{t+1}\) frame due to moving people. To mitigate this occlusion issue, our per-pixel minimum photometric loss is defined as follows,        
\begin{equation}
\begin{split}
    L_{photo} = \min_{I_{s}} L_{ph}(I_{t}, I^\prime_{t}).
\end{split}
\label{minimum_loss}
\end{equation}
Additionally, auto-masking encourages our method to mask out static pixels for which the appearance does not change between the current frame and temporally adjacent frames. This auto-masking results in removing pixels belonging to low-texture scenes (e.g. ceiling or fluorescent light) that have undesirable effects on the minimum photometric loss. Furthermore, these masks often remove non-Lambertian surfaces (e.g. mirrors or digital bilborad screens), which are common in the department store or the metro station. 

\subsubsection{Sparse Depth Loss}
Our proposed model can learn to encode absolute scale from sparse depth loss by aligning the depth prediction \(\hat{D}_{t}\) with sparse depth \(S_{t}\). We use the L1-norm to penalize the difference between \(\hat{D}_{t}\) and \(S_{t}\) over \(\Omega\), where sparse depth is available,      
\begin{equation}
\begin{split}
    L_{depth} = \sum_{x \in \Omega}|\hat{D}_{t}(x) - S_{t}(x)| \:.
\end{split}
\label{depth_loss}
\end{equation}

\subsubsection{Total Loss Function}
Following \cite{2017Godard,2019ICRAMa,Monodepth2}, we also incorporate an edge-aware smoothness loss function \(L_{smooth}\) over the predicted depth map to regularize the depth in texture-less low-image gradient regions. 
\begin{equation}
    L_{smooth} = {\mid\partial_{x}d^{*}\mid}e^{-{\mid\partial_{x}I_{t}\mid}} + {\mid\partial_{y}d^{*}\mid}e^{-{\mid\partial_{y}I_{t}\mid}},
\label{smooth_loss}
\end{equation}
where \(\partial_{x}, \partial_{y}\) denotes the gradients along the either the x or y direction. \(d^{*} = \hat{d}_{t} / {\overline{\hat{d}_{t}}}\) is the normalized inverse depth \cite{2018DDVO}. 
Our overall loss function is a weighted sum over the aforementioned losses,
\begin{equation}
    L_{tot} = L_{photo} + \lambda_{d}L_{depth} + \lambda_{s}L_{smooth}
\label{total_loss}
\end{equation}
where \(\lambda_{d}\) and \(\lambda_{s}\) denote weighting terms selected through a grid search.

\begin{figure}[t]
\centering
\includegraphics[width=0.9\linewidth]{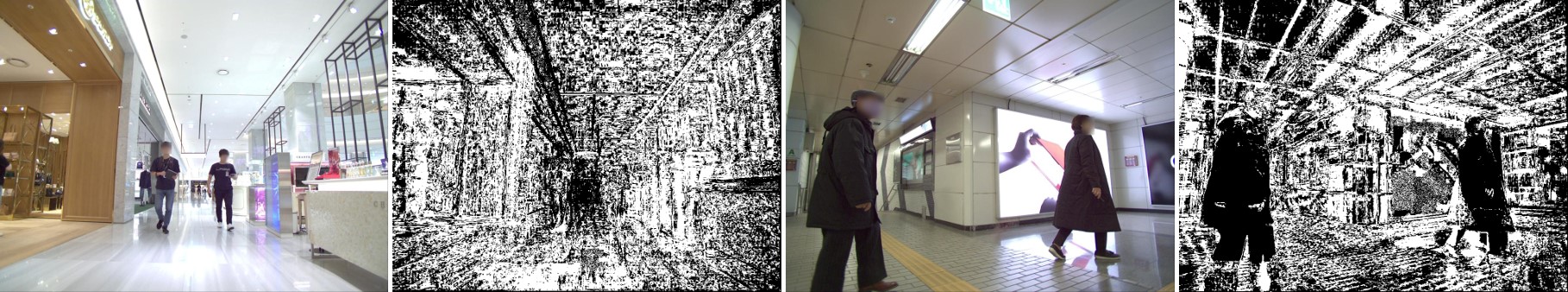}
\vspace{-3mm}
\caption{Examples of auto-masks computed after one epoch. Auto masks (black pixels) usually contain low-texture scenes or non-Lambertain surfaces where photometric loss is unreliable.}
\label{figure:auto-masks}
\vspace{-4mm}
\end{figure}

\subsection{Network Architecture}
In our training framework, the objective function defines a trade-off between a sparse depth loss, which enables our network to predict absolute scale depth on sparse regions, and a photometric loss, which encourages the network to predict dense depth where sparse information does not exist. In challenging indoor environments, self-supervised methods with photometric loss have shown that finding the optimal depth value is very difficult because the loss only comes from the appearance difference between target and synthesized images; measuring these differences is difficult due to the large non-texture regions, repeating structures, and occlusions.

The sparse depth loss might easily overwhelm the whole self-supervised training, and the depth predictions often result in overfitting on sparse depth input. Both feature extraction from the sparse depth and feature fusion schemes of multi-modal data are critical to avoid such negative effects. Our network adopts a late fusion scheme that uses each encoder for different sources because image and sparse depth features are heterogeneous. Throughout several experiments in Sec. \ref{subsubsection:naverlabs}, a late fusion scheme is shown to be particularly suitable for self-supervised depth completion. We focus on designing an effective architecture that leverages sparse depth features to fill the empty regions rather than getting stuck producing sparse depth prediction.   


\begin{figure}[t]
\centering
\includegraphics[width=0.9\linewidth]{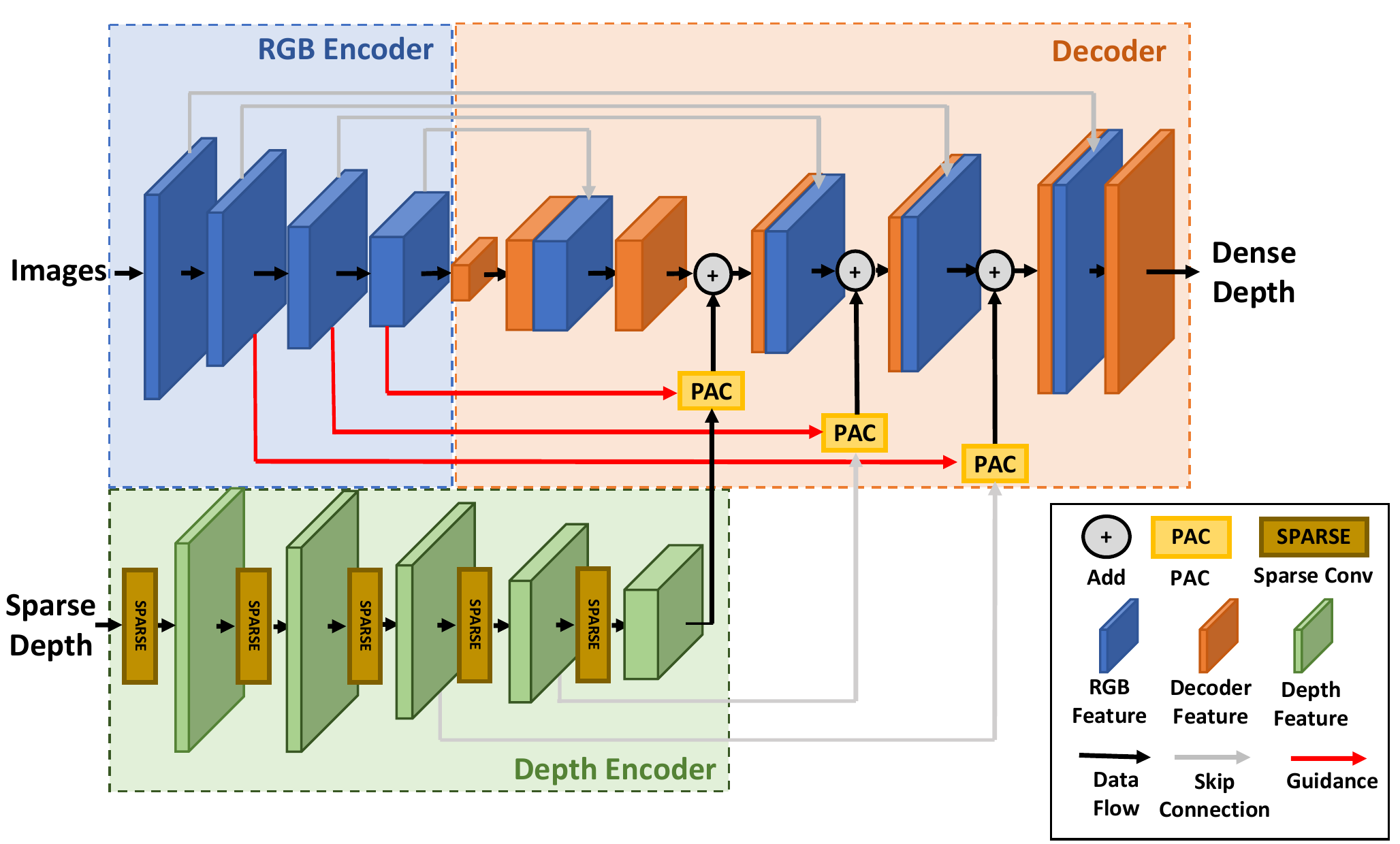}
\caption{ Our model consists of three main components: (1) an RGB encoder that extracts RGB features, reducing the spatial resolution of the input images to consider the global context; (2) a depth encoder that extracts sparse depth features from sparse input via sparse convolutional blocks; (3) a decoder that takes both RGB and depth features, and transforms them into dense depth maps. Data flow contains convolution operations. 
}
\label{figure2}
\vspace{-6mm}
\end{figure}

\subsubsection{Sparse Convolution}
In order to extract features from sparse depth, recent depth completion methods fall into two groups. One group \cite{SparsityCNN,2018Eldesokey} designs shallow networks with sparse convolution layers due to the poor performance of standard convolution layers on sparse input. In contrast, the other group \cite{2018ECCVcspn,2019ICRAMa,CSPN++} demonstrates that employing standard convolution layers yields better performance with their training schemes; these methods also perform well on different levels of sparsity. Comparing both approaches, we conclude that deep stacks of standard convolution layers 
are key to handling sparse depth input. Therefore, we choose to combine these two approaches by stacking sparse convolution layers \cite{SparsityCNN} deeper. Sparse convolution blocks take a sparse depth feature and a binary observation mask as inputs, which both have the same spatial size. Since an observation mask indicates the reliability of each location of the feature maps, the convolution operations only attend on reliable input features which are involved with sparse depth input. As a result, the depth encoder is able to propagate depth information from reliable sparse depth input to its surrounding depth features. 
In contrast to sparsity invariant convolutions \cite{SparsityCNN}, we do not apply mask normalization, which causes a degradation problem due to the small values extracted from the depth encoder when stacking deep layers.   

\subsubsection{Pixel-Adaptive Convolution (PAC)}
Recent works utilize operations like concatenation or element-wise addition to fuse the information from both the RGB and the depth encoder. However, this simple method is not the optimal way to fuse the information from two modalities because extracted depth features are not guided by RGB images in a late-fusion manner. To achieve this, we require spatially-variant convolution weights that different convolution kernels depending on image features are applied to different spatial positions of the depth features. Thus, we adopt pixel-adaptive convolutions (PACs) introduced by Su et al. \cite{2019CVPRpacnet}. Assume that the PACs take a sparse depth feature map \(\textbf{v}\) and an image feature \(\textbf{f}\) as input, and produce the output sparse depth feature \(\textbf{v}^\prime\). We denote the convolution weight as \(\textbf{W}\) and the bias term as \(\textbf{b}\). PACs augment the spatially invariant convolution weight \(\textbf{W}\) with a local-adaptive kernel function \(\textbf{K}\). Following the notation in \cite{2019CVPRpacnet}, the PAC convolution is defined as
\begin{equation}
\begin{split}
    \textbf{v}^\prime_{i} = \sum_{j\in \Omega(i)} K(\textbf{f}_i, \textbf{f}_j) \textbf{W}[\textbf{p}_i - \textbf{p}_j]\textbf{v}_j + \textbf{b}
\end{split}
\label{photometric_loss}
\end{equation}
where \(\textbf{f}\) are image features from the RGB encoder and act as guidance for the PACs in the depth encoder. High-level image features are preferable because they contain image context information. The kernel function \(\textbf{K}\) has a standard Gaussian form: \(K(\textbf{f}_i, \textbf{f}_j) = exp(-\frac{1}{2}(\textbf{f}_i - \textbf{f}_j)^T(\textbf{f}_i - \textbf{f}_j))\). This kernel computes the correlation between image features and guides the standard convolutional weights to generate depth features depending on the content learned from the RGB encoder. This spatially-variant kernel is helpful because, for instance, depth features on a person should not be applied to generate depth maps on the wall. Thus, PACs enable our proposed architecture to generate intermediate depth features consistent with image content.  





\subsubsection{Decoder}
The decoder is similar to Ma et al \cite{2019ICRAMa} with one major difference. Unlike their method, our proposed network predicts inverse depth instead of depth. Thus, we replace the ReLU nonlinearity function with sigmoid activation. We observe that applying a sigmoid function is critical for training stability and is more robust to diverse depth ranges on challenging indoor scenes. Without inverse depth prediction, we fail to train the network proposed by \cite{2019ICRAMa} on our challenging indoor datasets. 

\section{Implementation and Performance}
\subsection{Datasets} \label{subsection:Datasets}
To demonstrate that our proposed method performs robustly in the challenging environment, we experiment with NAVERLABS indoor dataset. Additionally, we evaluate our method on two benchmarks: the NYUv2 dataset and KITTI. For indoor scenes, we report five evaluation metrics: root mean squared error (RMSE), mean absolute relative error (ABS Rel), and $\delta_{i}$ ,which means the percentage of predicted pixels for which the relative error is less than a threshold \(i\). The RMSE is represented in \(mm\) scale. For the KITTI benchmark, we use four official metrics for evaluation.     

\subsubsection{NAVERLABS indoor dataset} \label{subsubsection:naverlabs}

NAVERLABS dataset
\cite{spoxelnet}  
comprises data collected from three different places: Department Store B1 (Dept. B1), Department Store 1F (Dept. 1F), and Metro Station (MS). All data were collected using a mapping robot equipped with two 16-channel LiDAR sensors and six RGB cameras with 2592\(\times\)2048 resolution. LiDAR SLAM is used to estimate sensor poses for our dataset, and this pose information enables us to generate depth maps. We captured 17K images for Dept. 1F, 32K images for Dept. B1, and 61K images for MS.  
The MS dataset was captured at a highly crowded metro station. 

\begin{figure}[t]
\centering
\includegraphics[width=0.9\linewidth]{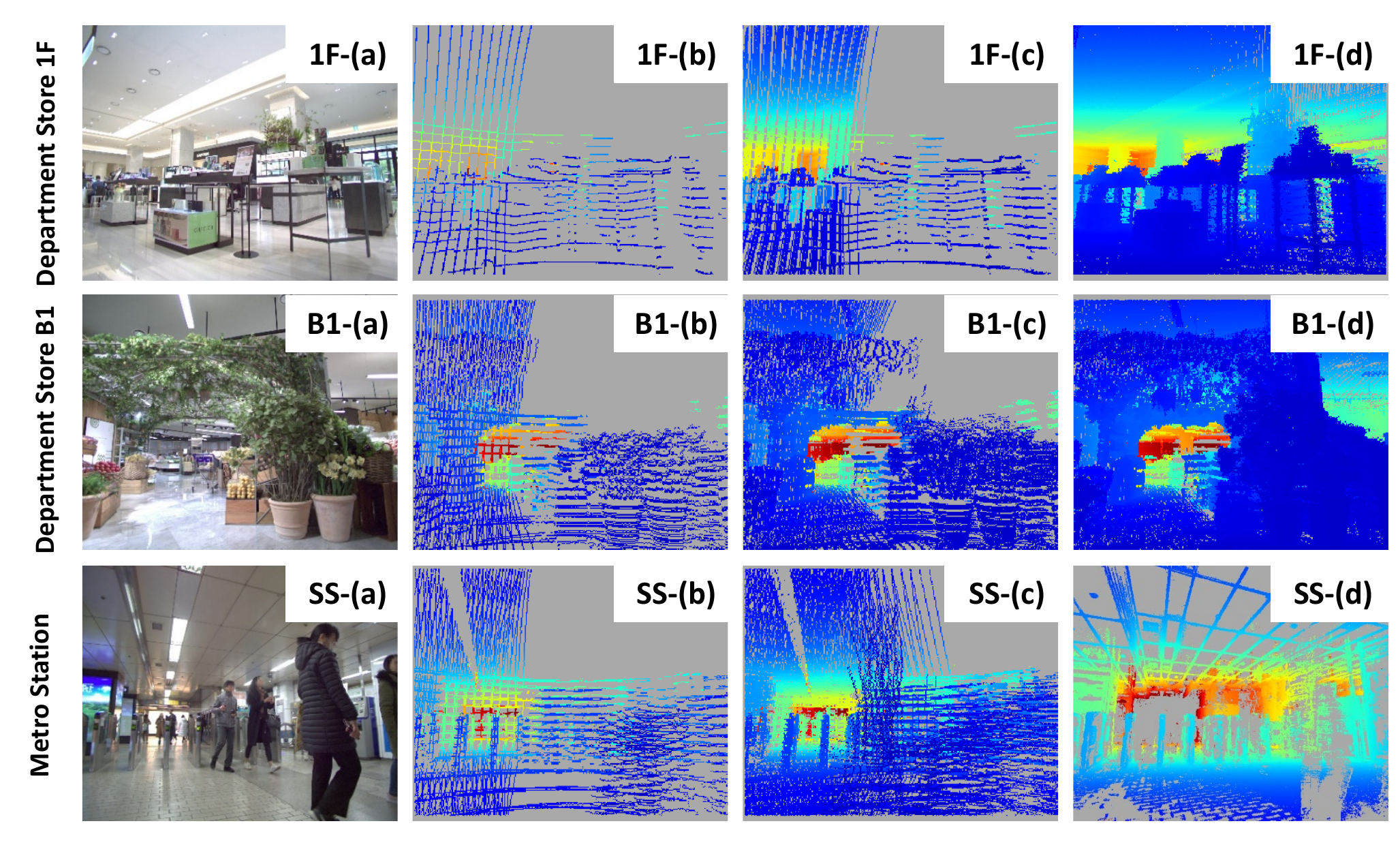}
\vspace{-3mm}
\caption{Samples from the NAVERLABS indoor dataset. We include more samples and demonstrate many challenging features from our datasets in supplementary video.
}
\vspace{-6mm}
\label{figure4}
\end{figure}

We accumulate consecutive LiDAR sweeps and project the point clouds onto each of six cameras in order to generate depth maps. These depth maps can be used as sparse input and ground truth for evaluation. In order to create sparse input, we accumulate 0.3 seconds of LiDAR scans, which correspond to one LiDAR scan since the LiDAR is sampled at 10 Hz (see the second column in Fig. \ref{figure4}). In the Dept. 1F and Dept. B1 datasets, we collect 1 second of LiDAR scans to generate ground truth depth for training supervised depth completion algorithms (see the third column in Fig. \ref{figure4}). However, the depth maps generated from the MS dataset are too noisy to serve as ground truth due to noise caused by moving objects (see SS-(c) in Fig. \ref{figure4}). For evaluation, we select the high-quality depth maps generated from projecting a LiDAR 3D model onto the camera frames in the Dept. 1F and Dept. B1 datasets (see 1F-(d) and B1-(d) in Fig. \ref{figure4}). In contrast, we use the SfM method \cite{SfM_revisited} to reconstruct a sparse 3D model of the MS dataset and MVS system \cite{MVS} to produce dense depth maps instead of noisy depth maps from LiDAR point clouds. 
Given the pose from LiDAR SLAM, we can obtain a 3D model aligned with the LiDAR data. SS-(d) in Fig. \ref{figure4} generated by the SfM is the example of depth maps for evaluation. 


\subsubsection{NYU-Depth-v2 dataset}
The NYUv2 dataset \cite{NYUv2} comprises RGB images and depth maps collected from Microsoft Kinect in 464 indoor scenes. The raw training dataset contains 268K images. We downsampled 10 times to remove redundant frames, resulting in 47k frames sampled for the training set. Our method is tested on the 654 official labeled test set for evaluation.

\subsubsection{KITTI depth completion dataset}
The KITTI depth completion dataset \cite{SparsityCNN} provides RGB images, sparse lidar points, and dense ground truth depth. Following the official split, it contains 86k frames for the training set, 1k frames for the validation set, and 1k frames for the evaluation.    

\subsection{Implementation Details}

We adopt ResNet34 \cite{ResNet} as the RGB encoder, and it was pretrained on ImageNet \cite{ImageNet}. 
The proposed deep networks were implemented with PyTorch \cite{PyTorch} and trained on 4 Tesla V100 GPUs. 
We trained our network from scratch for 30 epochs, with a batch size of 16 using Adam \cite{Adam}, where \(\beta_{1}=0.9\), \(\beta_{2}=0.999\); we used an input resolution of 1024\(\times\)760 for NAVERLABS, 640\(\times\)480 for NYUv2, and 1024\(\times\)320 for KITTI. We used an initial learning rate of \(10^{-4}\) for the first 10 epochs and halved it every 10 epochs. In loss functions, we set the \(\alpha\) to 0.85, \(\lambda_{d}\) to 0.001 and \(\lambda_{s}\) to 0.1.

\subsection{Experiments on NAVERLABS indoor datasets}

\begin{figure}[t]
\centering
\includegraphics[width=0.9\linewidth]{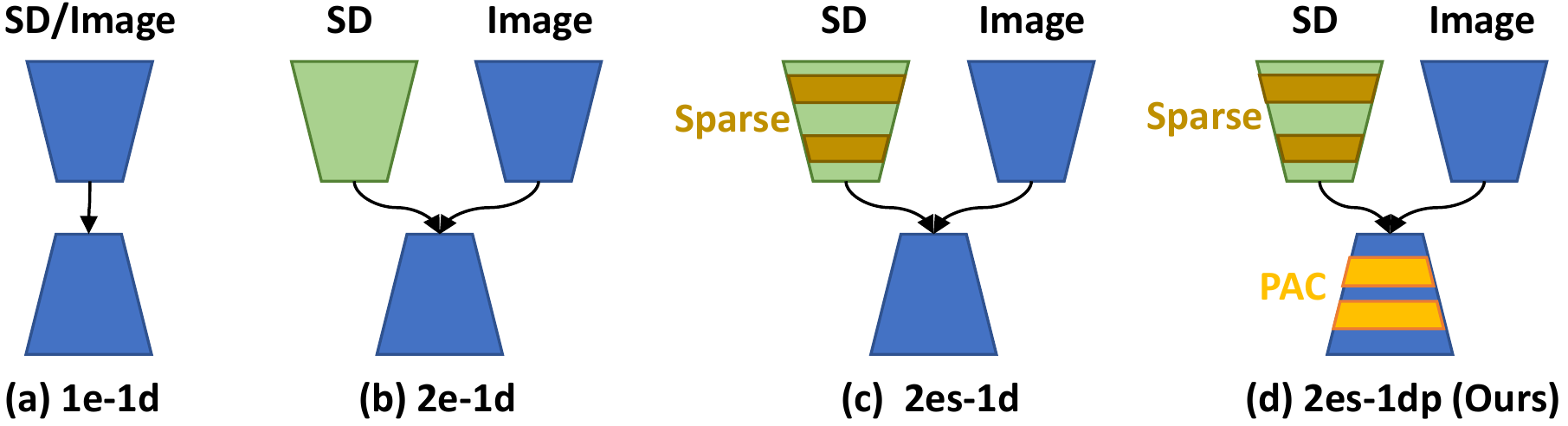}
\vspace{-3.2mm}
\caption{
SD represents a sparse depth. We design 4 architectures: (a) 1e-1d: an encoder and a decoder; (b) 2e-1d: an RGB encoder, a depth encoder, and a decoder; (c) 2es-1d: an RGB encoder, a depth encoder with sparse convolutions, and a decoder; (d) 2es-1dp: the same architecture as Fig. \ref{figure2}.  
}
\vspace{-2mm}
\label{fig:Ablation}
\end{figure}

\begin{table}[t] 
\begin{center}
\caption{\label{table:NAVERLABS} Depth completion results on NAVERLABS indoor dataset.}
\resizebox{0.48\textwidth}{!}{
\begin{tabular}{lcccccc}
\toprule
\multicolumn{7}{c}{(a)  Dept. 1F Dataset}\\
\midrule
Method & T & RMSE \(\downarrow\) & ABS Rel \(\downarrow\) & $\delta_{1.25}$ \(\uparrow\) & $\delta_{1.25^2}$ \(\uparrow\) & $\delta_{1.25^3}$ \(\uparrow\) \\
\midrule
UNET \cite{2019ICRAMa}& S & 5043.5 & 0.571 & 0.273 & 0.463 & 0.553 \\
nUNET \cite{2019Eldesokey}& S & 4993.8 & 0.171 & 0.898 & 0.933 & 0.951 \\
CSPN \cite{2018ECCVcspn}& S  & 2857.1 & 0.266 & 0.808 & 0.886 & 0.924 \\
MD2 \cite{Monodepth2}& SS  & 3862.4 & 0.493 & 0.488 & 0.716 & 0.818 \\
\midrule
1e-1d & SS  & 2899.6 & 0.256 & 0.784 & 0.869 & 0.911 \\
2e-1d & SS  & 2801.4 & 0.265 & 0.783 & 0.868 & 0.909 \\
2es-1d & SS  & 2700.3 & 0.242 & 0.790 & 0.875 & 0.915 \\
Ours & SS & \textbf{2692.9} & \textbf{0.234} & \textbf{0.815} & \textbf{0.887} & \textbf{0.926} \\
\bottomrule
\\
\toprule
\multicolumn{7}{c}{(b) Dept. B1 Dataset}\\
\midrule
Method & T & RMSE \(\downarrow\) & ABS Rel \(\downarrow\) & $\delta_{1.25}$ \(\uparrow\) & $\delta_{1.25^2}$ \(\uparrow\) & $\delta_{1.25^3}$ \(\uparrow\) \\
\midrule
UNET \cite{2019ICRAMa}& S & 3048.5 & 0.346 & 0.599 & 0.711 & 0.764 \\
nUNET \cite{2019Eldesokey}& S & 2963.1 & 0.324 & 0.635 & 0.732 & 0.779  \\
CSPN \cite{2018ECCVcspn}& S  & 2162.2 & 0.199 & 0.774 & 0.890 & 0.938  \\
MD2 \cite{Monodepth2}& SS  & 3592.7 & 0.338 & 0.47 & 0.729 & 0.862  \\
\midrule
1e-1d & SS  & 2210.5 & 0.186 & 0.823 & 0.914 & 0.947 \\
2e-1d & SS  & 2157.5 & 0.170 & 0.832 & 0.916 & 0.949 \\
2es-1d & SS  & 2116.8 & 0.183 & 0.829 & 0.916 & 0.949\\
Ours & SS & \textbf{2024.1} & \textbf{0.178} & \textbf{0.841} & \textbf{0.922} & \textbf{0.953} \\
\bottomrule
\\
\toprule
\multicolumn{7}{c}{(c) Metro Station Dataset}\\
\midrule
Method & T & RMSE \(\downarrow\) & ABS Rel \(\downarrow\) & $\delta_{1.25}$ \(\uparrow\) & $\delta_{1.25^2}$ \(\uparrow\) & $\delta_{1.25^3}$ \(\uparrow\) \\
\midrule
UNET \cite{2019ICRAMa}& S & 4029.7 & 0.749 & 0.201 & 0.219 & 0.23  \\
nUNET \cite{2019Eldesokey}& S & 2705.1 & 0.472 & 0.444 & 0.528 & 0.582 \\
MD2 \cite{Monodepth2}& SS  & 1321.2 & 0.166 & 0.8 & 0.916 & 0.939 \\
Ours & SS & \textbf{1248.4} & \textbf{0.138} & \textbf{0.866} & \textbf{0.923} & \textbf{0.942} \\
\bottomrule
\end{tabular}
}
\end{center}
``S'' represents \textit{supervised} training methods and ``SS'' denotes \textit{self-supervised} training methods. Different from other depth completion methods, MD2 is based on self-supervised depth estimation with monocular video sequences. Compared to prior self- and fully- supervised methods, our proposed method demonstrates improvement in RMSE metric by 5.8\% in Dept. 1F, by 6.4\% in Dept. B1 and by 5.5\% in Metro Station. 
\vspace{-6mm}
\end{table}

\begin{figure}[t]
\centering
\includegraphics[width=0.9\linewidth]{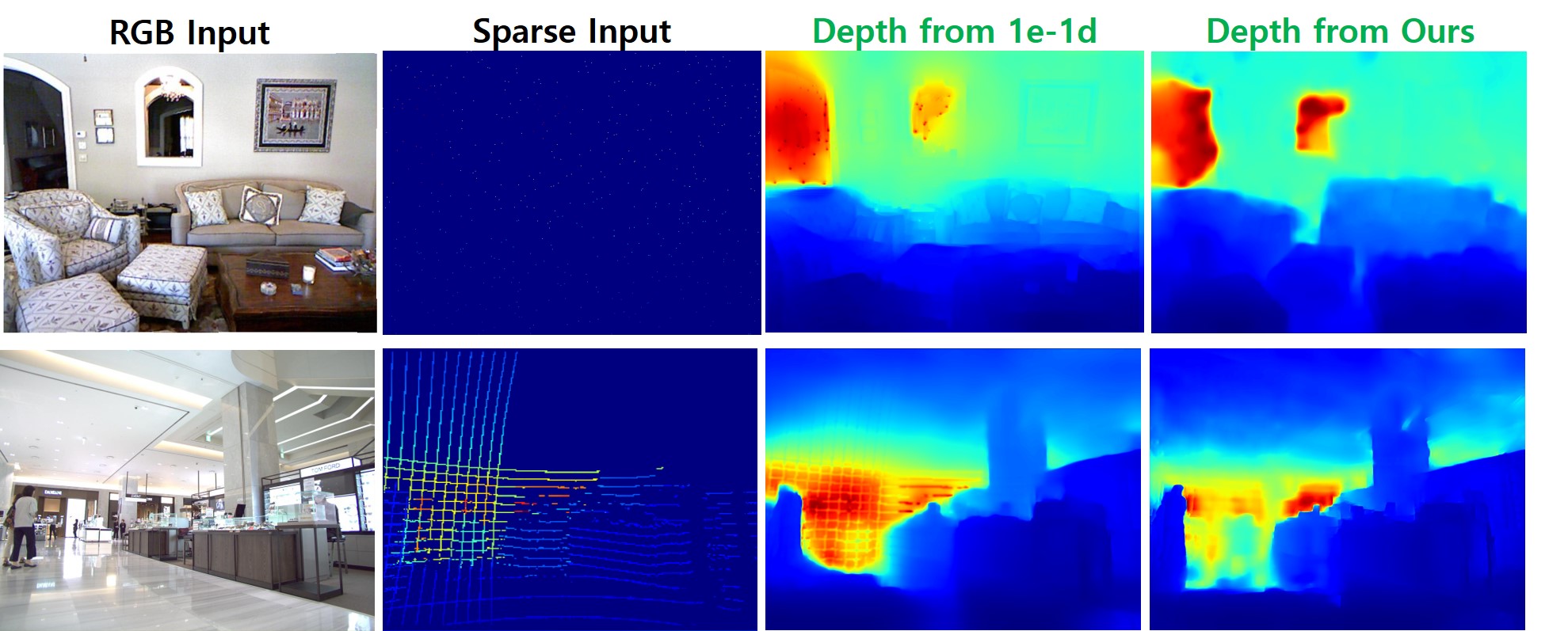}
\vspace{-2mm}
\caption{Examples of depth artifacts generated by 1e-1d model. The top (NYUv2) and bottom (Dept. 1F) figures in third column shows irregular sparse patterns copied from sparse inputs and blurred depth boundaries}
\label{figure:Analysis}
\vspace{-6mm}
\end{figure}

\begin{figure*}[t]
    \centering
    \includegraphics[width=0.85\linewidth]{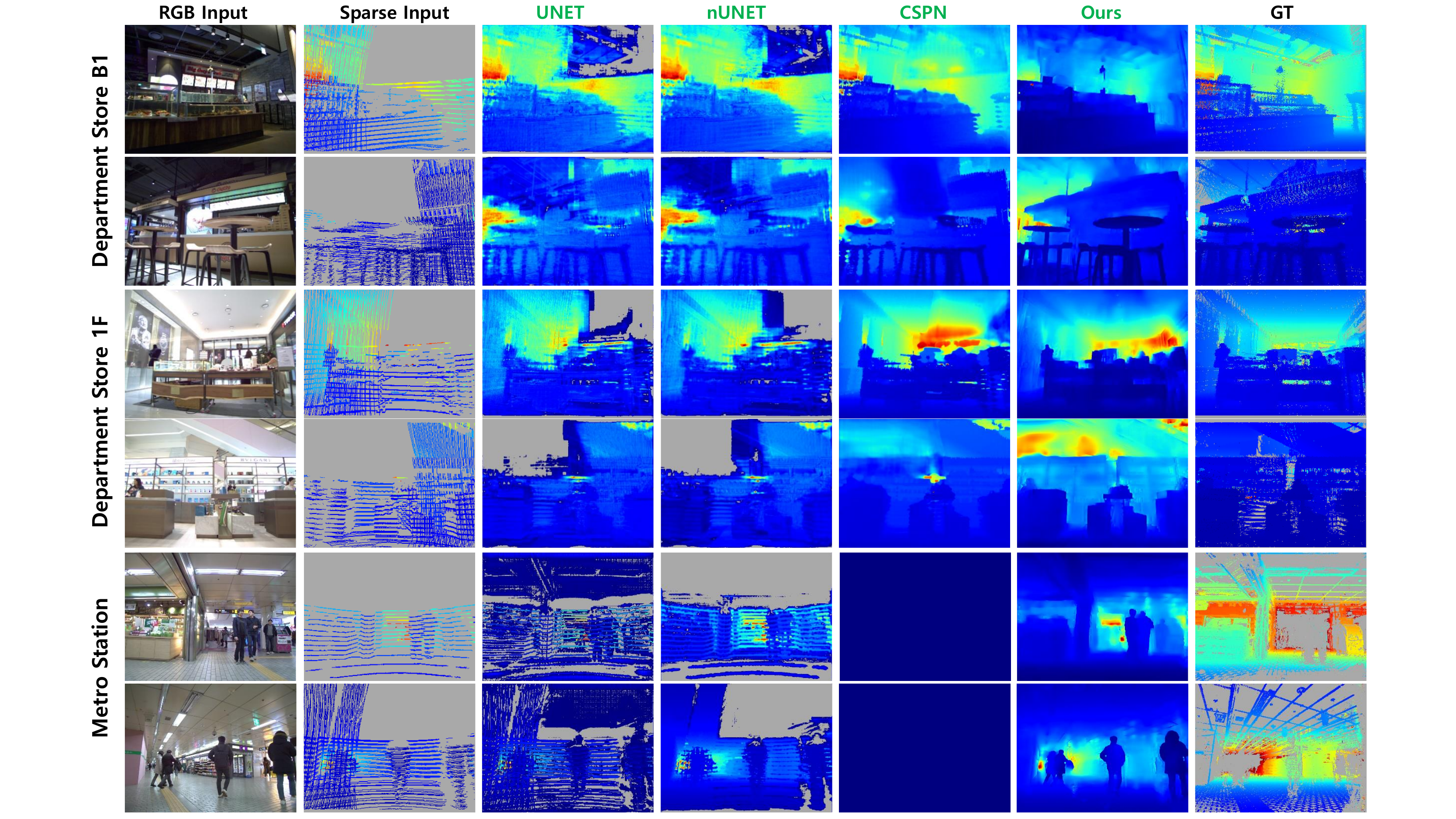}
    \vspace{-2mm}
    \caption{Qualitative comparison with ``UNET''  \cite{2019ICRAMa}, ``nUNET''  \cite{2019Eldesokey}, and ``CSPN'' \cite{2018ECCVcspn} on the NAVERLABS indoor dataset. 
    As mentioned in Sec. \ref{subsubsection:naverlabs}, the groundtruth of the Metro Station dataset is the results of MVS. 
    (low \protect\includegraphics[width=1cm,height=0.2cm]{ICRA_figures/colorbar.jpeg} high; grey means empty depth values.)
    }
    \label{fig:comparison}
    \vspace{-4mm}
\end{figure*}

We compare our proposed method with 4 baseline approaches with public codes: 1) UNET is a U-shaped ResNet \cite{2019ICRAMa}; 2) nUNET is an early-fusion encoder-decoder network combined with multiple normalization convolutions \cite{2019Eldesokey}; 3) CSPN learns the affinity to perform sparse depth propagation \cite{2018ECCVcspn}; 4) Monodepth2 (MD2) is based on ResNet34 \cite{Monodepth2}. The same relative pose generated from our method is applied to training MD2 for a fair comparison. MD2 benefits from the use of ground truth median scaling at inference, while other depth completion methods utilize unmodified network prediction. Except for 4), the aforementioned methods are based on supervised training with ground truth depth. As mentioned in Section \ref{subsection:Datasets}, it is difficult to create ground truth depth for the Metro Station dataset (MS). In the case of the MS dataset, we only use the sparse depth maps as weak supervisions for supervised depth completion. Quantitative results are shown in Table \ref{table:NAVERLABS}, and the associated qualitative comparisons are shown in Fig. \ref{fig:comparison}. We train the model from Ma et al. \cite{2019ICRAMa} with its network architecture and the relative poses from our method, which are more accurate. Note that Ma et al. failed to produce meaningful results. We observe that the network just copies the sparse inputs, meaning that the sparse depth loss only serves as supervision for training while the photometric loss is unreliable. 

In contrast, our method successfully performs depth completion in the NAVLERLABS dataset. This indicates that our training framework is more robust to variant supervision compared to other self-supervised depth completion methods. The performance of our method is superior to existing supervised completion methods. From columns 4 through 6 of Fig. \ref{fig:comparison}, we can observe that supervised depth completion methods rely heavily on the density of ground truth depth. We notice that these algorithms fail to fill the holes where LiDAR information does not exist (e.g., the upper part of the image) due to the sensor configuration. In contrast, thanks to the effectiveness of photometric loss, the proposed method succeed in filling empty regions such as the ceiling, which LiDAR sensors are not able to measure, with accurate depth values. In addition, our method even predicts the correct depth map on shiny or transparent surface which cannot be scanned accurately by sensors (see First row in Fig. \ref{fig:comparison}). The experimental results of CSPN bolster this observation because our method achieves performance comparable to the supervised setting in Table \ref{table:NAVERLABS}-(a), (b), while its training with weak supervision collapsed on the MS dataset. If we collect the high-quality dense depth maps (see the last column in Fig. \ref{fig:comparison}), the supervised depth completion methods can be superior to our method. However, collecting dense depth maps requires a high cost and can be impractical in certain environments such as the MS dataset.              


\begin{table}[t]
\begin{center}
\caption{\label{table:NYUv2} depth completion results on the NYUv2 dataset}
\resizebox{0.48\textwidth}{!}{
\begin{tabular}{lcccccc}
\toprule
Method & T & RMSE \(\downarrow\) & ABS Rel \(\downarrow\) & $\delta_{1.25}$ \(\uparrow\) & $\delta_{1.25^2}$ \(\uparrow\) & $\delta_{1.25^3}$ \(\uparrow\) \\
\midrule
\multicolumn{7}{l}{(a) 500 Samples}\\
Ma et al. \cite{2018ICRAMa} & S & 0.204 & 0.043 & 97.8 & 99.6 & 99.9 \\
CSPN \cite{2018ECCVcspn} & S & 0.117 & 0.016 & 99.2 & 99.9 & 100 \\
Ma et al. \cite{2019ICRAMa} & SS & 0.271 & 0.068 & - & - & - \\
Ours & SS & 0.178 & 0.033  & 98.1 & 99.7 & 100 \\
\midrule
\multicolumn{7}{l}{(b) 200 Samples}\\
Ma et al. \cite{2018ICRAMa} & S & 0.230 & 0.044 & 97.1 & 99.4 & 99.8 \\
Yang et al. \cite{2019CVPRYang} & SS & 0.569 & 0.171  & - & - & - \\
Yoon et al. \cite{2020IROSYoon} & SS & 0.309 & -  & - & - & - \\
Ours  & SS & 0.240 & 0.048 & 96.6 & 99.4 & 99.9 \\
\bottomrule
\end{tabular}
\vspace{-3mm}
}

\caption{\label{table:KITTI} depth completion results on the KITTI validation set}
\resizebox{0.48\textwidth}{!}{
\begin{tabular}{lccccc}
\toprule
Method & T & RMSE [mm] & MAE [mm] & iRMSE [1/km] & iMAE [1/km] \\
\midrule
Uhrig et al. \cite{SparsityCNN} & S & 1601.33 & 481.27 & 4.94 & 1.78  \\
Ma et al. \cite{2019ICRAMa} & S & 814.73 & 249.95 & 2.80 & 1.21  \\
\midrule
Yoon et al. \cite{2020IROSYoon} & SS & 1593.37 & 547.00 & 27.98 & 2.36 \\
Ma et al. \cite{2019ICRAMa} & SS & 1343.33 & 358.66 & 4.28 & 1.64  \\
Yang et al. \cite{2019CVPRYang} & SS & 1310.03 & 347.17 & - & - \\
Ours & SS & 1212.89 & 346.12 & 3.54 & 1.29 \\
\bottomrule
\end{tabular}
}

\end{center}
``S'' represents \textit{supervised} training methods and ``SS'' denotes \textit{self-supervised} training methods. 
Compared to prior self-supervised method, our method reduces the RMSE by 34.3\% in NYUv2 and by 7.4\% in KITTI. 
\vspace{-6mm}
\end{table}

To investigate the impact of our network architecture, we conduct ablation studies for different network configurations on the NAVERLABS dataset. Figure \ref{fig:Ablation} illustrates the network variants, and their quantitative results are shown in Table \ref{table:NAVERLABS}. Comparing the results of 1e-1d with those of 2e-1d, we observe that fusing the RGB and depth feature at the earliest stage worsens the results. With the early fusion, we often observe depth with blurrier depth discontinuities, as seen in Fig. \ref{figure:Analysis}. With PAC, depth maps can eliminate either irregular lattice patterns from LiDAR or sparse points from sampled depth maps. We conclude that the late fusion is more effective for self-supervised learning because image and depth features are heterogeneous data, and early fusion is more likely to produce noisy initial depth values.   


\subsection{Experiments on Public Benchmark Datasets}

To verify the performance of our method on public datasets, we train and evaluate our method on the NYUv2 dataset. Following previous works, we uniformly sample 200 and 500 sparse depth points separately for sparse input. Except for supervised depth completion methods, our proposed method achieves the best performance among self-supervised depth completion methods in Table \ref{table:NYUv2}.


In order to prove the generalization capability of our method, we train and evaluate our method on an outdoor dataset, the KITTI depth completion dataset. The comparison results are shown in Table \ref{table:KITTI}. We observe that our method outperforms other self-supervised depth completion methods on every metric. Although Yang \cite{2019CVPRYang} pretrains their network using a virtual dataset such as the Virtual KITTI dataset \cite{VKITTI}, which is similar to the benchmark dataset, our method performs better than their method. We include qualitative results in the supplementary video.          



\section{Conclusions, Limitations, and Future Work}
In this paper, we investigated the self-supervised depth completion framework for a challenging indoor environment. We provide two main research contributions. First, we introduced a training framework that leads to stable self-supervised training. Second, we proposed a novel architecture to combine image features and sparse depth features effectively. In indoor environments, our method outperforms other self-supervised methods and is competitive with supervised methods. Furthermore, it generalizes better to all public benchmark datasets. As part of future work, we would like to apply this method to complement depth sensors for mobile robots in indoor environments and and evaluate its benefits for robot navigation. A limitation of the current method is the high computational cost for embedded platforms. For robotic tasks, we would like to explore network compression algorithms to reduce computational complexity and latency.  

\addtolength{\textheight}{0cm}




\section*{ACKNOWLEDGMENT}
This work was supported by the Institute of Information \& Communications Technology Planning \& Evaluation(IITP) grant funded by the Korea government(MSIT) (No. 2019-0-01309, Development of AI Technology for Guidance of a Mobile Robot to its Goal with Uncertain Maps in Indoor/Outdoor Environments). This work was supported in part by ARO Grants W911NF1910069, W911NF2110026  and W911NF1910315 and NSF grant 2031901.


\end{document}